\documentclass[sigconf]{acmart}
\usepackage{amsmath}

\DeclareMathOperator*{\argmin}{arg\,min}
\usepackage{amsfonts}
\usepackage{booktabs}
\usepackage{hyperref}
\usepackage{graphicx}
\usepackage{multirow}

\begin{document}
%
% --- Author Metadata here ---
% -- Can be completely blank or contain 'commented' information like this...
%\conferenceinfo{WOODSTOCK}{'97 El Paso, Texas USA} % If you happen to know the conference location etc.
%\CopyrightYear{2001} % Allows a non-default  copyright year  to be 'entered' - IF NEED BE.
%\crdata{0-12345-67-8/90/01}  % Allows non-default copyright data to be 'entered' - IF NEED BE.
% --- End of author Metadata ---

\title{Improving Native Ads CTR Prediction by Large Scale Event Embedding and Recurrent Networks}
%\subtitle{[Extended Abstract]
% You need the command \numberofauthors to handle the "boxing"
% and alignment of the authors under the title, and to add
% a section for authors number 4 through n.
%
% Up to the first three authors are aligned under the title;
% use the \alignauthor commands below to handle those names
% and affiliations. Add names, affiliations, addresses for
% additional authors as the argument to \additionalauthors;
% these will be set for you without further effort on your
% part as the last section in the body of your article BEFORE
% References or any Appendices.

%\numberofauthors{5}
%
% You can go ahead and credit authors number 4+ here;
% their names will appear in a section called
% "Additional Authors" just before the Appendices
% (if there are any) or Bibliography (if there
% aren't)

% Put no more than the first THREE authors in the \author command
%%You are free to format the authors in alternate ways if you have more 
%%than three authors.

%\author{
%
% The command \alignauthor (no cURLy braces needed) should
%% e-mail address with \email.
%\subtitle{Extended Abstract}
%\subtitlenote{The full version of the author's %guide is available as
%  \texttt{acmart.pdf} document}

\author{Mehul Parsana}
\affiliation{%
  \institution{Microsoft Corporation}
  \streetaddress{}
  \city{Redmond}
  \state{WA}
  \postcode{}
}
\email{mparsana@microsoft.com}

\author{Krishna Poola}
%\authornote{}
\affiliation{%
  \institution{Microsoft Corporation}
  \streetaddress{}
  \city{Redmond}
  \state{WA}
  \postcode{}
}
\email{krishnap@microsoft.com}

\author{Yajun Wang}
%\authornote{This author is the
%  one who did all the really hard work.}
\affiliation{%
  \institution{Microsoft Corporation}
  \streetaddress{}
  \city{Sunnyvale}
  \state{CA}
}
\email{yajunw@microsoft.com}

\author{Zhiguang Wang}
\authornote{The authors are listed in alphabetical order.}
\affiliation{%
  \institution{Microsoft Corporation}
  \city{Redmond}
   \state{WA}
}
\email{zhigwang@microsoft.com}

%\additionalauthors{Additional authors: John Smith (The Th{\o}rvald Group,
%email: {\texttt{jsmith@affiliation.org}}) and Julius P.~Kumquat
%(The Kumquat Consortium, email: {\texttt{jpkumquat@consortium.net}}).}

\begin{abstract}
Click through rate (CTR) prediction is very important for Native advertisement but also hard as there is no direct query intent. In this paper we propose a large-scale event embedding scheme to encode the each user browsing event by training a Siamese network with weak supervision on the users' consecutive events. The CTR prediction problem is modeled as a supervised recurrent neural network, which naturally model the user history as a sequence of events. Our proposed recurrent models utilizing pretrained event embedding vectors and an attention layer to model the user history. Our experiments demonstrate that our model significantly outperforms the baseline and some variants.
\end{abstract}

\maketitle
\section{Introduction}
Native advertisement is a new type of online advertisement which attracts significant attention in recent years, especially due to the effectiveness of Facebook ads~\cite{facebookads}, Yahoo Gemini ~\cite{yahoogemini} and Bing Intent Ads~\cite{bingintent}. In native ads scenarios, the advertisements replicate the look and feel of the organic contents and are placed interchangeably with them. It has been very successful in providing good user experience and effective return of investment for advertisers~\cite{sharethrough}.

One distinctive characteristic of native ads, comparing with traditional search ads, is the lack of strong user query intent. In search ads, the search query from user input plays a dominant role in optimizing the serving system. In native ads, however, it is important to infer users' intent from their history of online activities. 

We propose an event embedding scheme to map the events from users' browsing activities we collected from our advertisers  to a latent space. The browsing history consists of URLs the user visited as well as titles and simple descriptions. With our event embedding scheme, the sparse activity data is embedded into a fixed length vector with weak supervision by the user browsing history only. 
%In particular, the embedding is trained outside any click and query data, which can be plugged in any other models.

We apply our event embedding scheme to the native ads click through rate prediction problem. 
In each request, for each advertisement that is selected, we need to compute the click through rate, i.e., how likely the advertisement will be clicked by the user in this request. The CTR estimation is very critical in selecting and ranking the advertisements as well as downstream optimization processes. We employ recurrent neural network models to predict the CTR from users' browsing activities. We show that by applying the event embedding with our designed recurrent attentional model, we can achieve the performance that is significantly better than the baseline and other variants.

\section{Related Work}
%{\bf DNN for recommendation.}

Deep neural networks has been applied in personalized recommendation systems.
He et~al~\cite{he2017neural} combine neural networks and matrix factorization machine to model user-item interactions. They model the user's preference implicitly by generating the latent vector jointly with the item matrix with a supervised loss. Long short-term memory (LSTM) based sequence modeling is applied in~\cite{okura2017embedding}
for news recommendation. They generates user representations with an
recurrent neural networks from user news browsing histories. Their embedding layer consists of a denoising autoencoder~\cite{Vincent:2008:ECR:1390156.1390294}. The embedding layer is not applicable in our setting as we do not have rich content information for the users' past browsing activities. Deep interest network (DIN)~\cite{DIN} is proposed to estimate click through rates for advertisements in e-commence site. Their event embedding layer is trained together with the user modeling. Also their model is not sequential based, as their model cannot learn from long term history. Recurrent Neural Network is used in~\cite{zhang2014sequential} for click prediction in search ads scenario with only users search history.

There are limited work in user browsing event embedding especially in the context of modeling user history. Our approach is inspired by the word2vec~\cite{word2vec} and the metric learning with Siamese networks ~\cite{chopra2005learning}, which is successfully applied in a lot of natural language processing and computer vision tasks. In particular, the word vector is usually trained outside of the actual NLP tasks. The framework of word embedding is also utilized to train the item vector in achieving good performance in genre classification for Windows 10 applications~\cite{barkan2016item2vec}.
Their embedding is ID based, thus unable to handle unseen items. The model also discards the temporal information which is very important in our setting. A character level CNN is proposed to learn a URL embedding in to detect malicious URLs \cite{saxe2017expose}. It trains the embedding using a binary classifier specific to the task and hence is not applicable in our problem. DSSM is proposed in ~\cite{huang2013learning} to train the word embedding through click on the query-document pairs, which usually requires huge number of labeled data.

Piece-wise Linear Models (LS-PLM) ~\cite{gai2017learning} and factorization machine (FM) ~\cite{rendle2010factorization} models can be viewed as a class of networks with one hidden layer, which first employs embedding layer on sparse inputs and then imposes special designed transformation functions for output, aiming to capture the interactive relationships among features. As the scale of feature and sample becoming larger and larger, the CTR prediction model has evolved from shallow to deep structure in recent years. Wide$\&$Deep ~\cite{cheng2016wide} and the YouTube Recommendation CTR model ~\cite{covington2016deep} extend the idea of factorization machine by replacing the transformation function with feed forward networks, which improve the model capability. They follow a similar model structure that stacking an embedding layer before fully connected layer with pooling layer to integrate the multi-dimension hidden vectors, which is now more likely to be the de facto standard. Our baseline model follows this kind of model structure.

\section{Latent Event Representation} 
\label{sec:event_vector}
As mentioned earlier, one crucial component in our models is an event embedding scheme that embeds each user browsing event into a fixed dimensional space. In this section, we describe the scheme in detail.

For each user browsing activity $e_u^t$ from user $u$ at time $t$, we define it as a triplet 

$$ e_u^t =\{U, T, D\},$$
where $U$ is the URL, $T$ is the title of the URL page and $D$ is a simple description where we received from our advertiser websites. A user session $S = \{e_1, e_2, e_k\}$ is defined as a sequence of events the user visits in a session. \footnote{In practice, we take all consecutive events that are within $30$ minutes apart as a session. }
We train a Siamese network \cite{chopra2005learning} to embed events $\{e_t\}$ into a low dimensional latent space.

\begin{figure*}[t]
	\centering
	\includegraphics[width=0.7\textwidth]{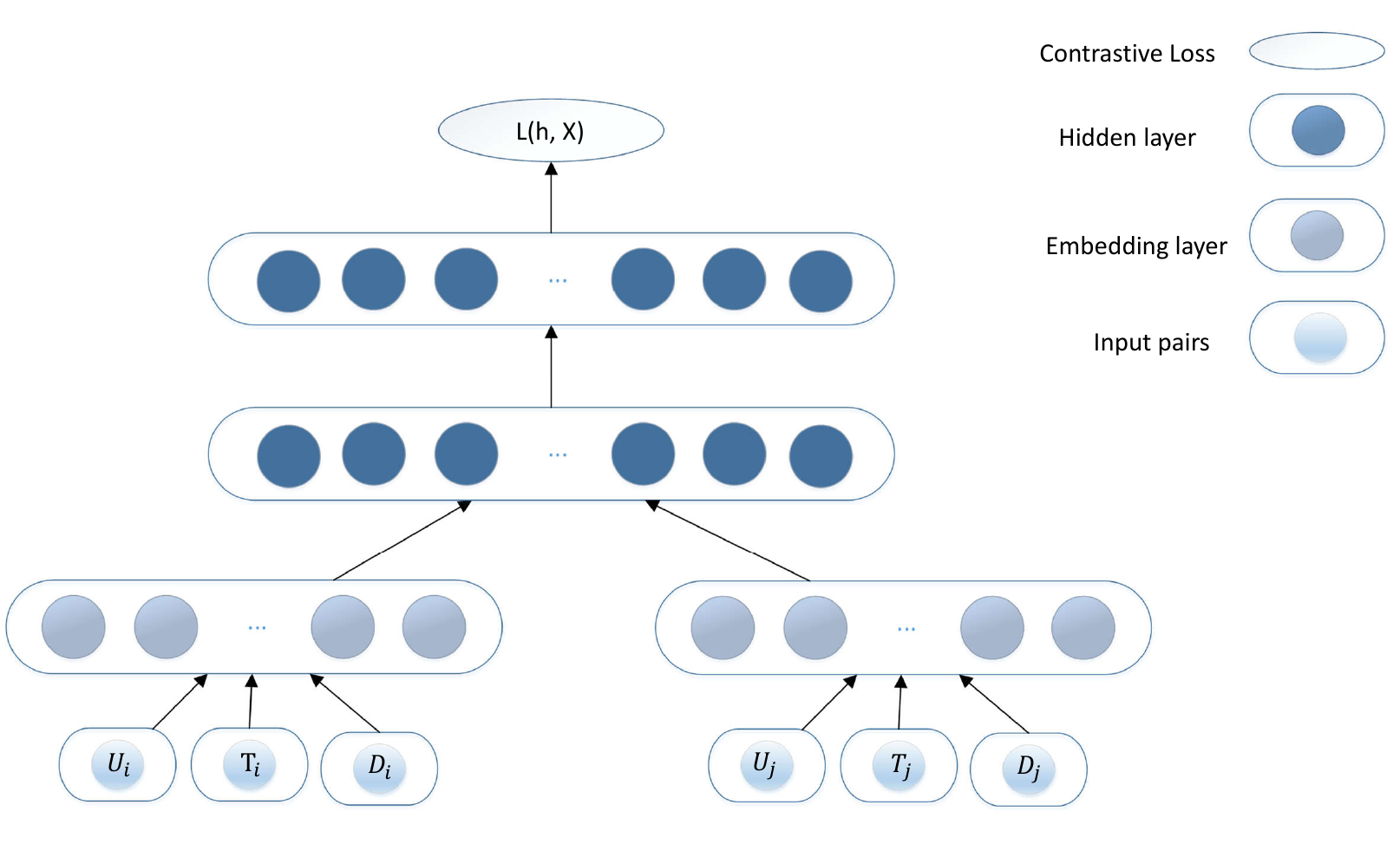}
	\caption {Siamese network structure to learn the event embedding.}
	\label{fig:Siamese}
\end{figure*}

\subsection{Network Structure}
The Siamese networks learn the latent embeddings of the user events by enforcing the distance of two consecutive events to be close. The input are events $\{e_i, e_j\} \in S$ which are visited by one user consecutively in a session $S$. We employ negative sampling on random event pairs $\{\tilde{e_i}, \tilde{e_i}\}$ to get the same number of negative samples. 

The Siamese networks consists of two identical neural networks, each taking one of the two input events. The last layers of the two networks are then fed to a contrastive loss function , which calculates the similarity between the two events.

Our networks is formulated as:
\begin{eqnarray}
\tilde{X_i} &\sim& I(e_i) \qquad   e_i = \{U_i, T_i, D_i\} \nonumber \\
\tilde{X_j} &\sim& I(e_j) \nonumber \\
h_i &=&  f(W\tilde{X_i} +b) \nonumber \\
h_j &=&  f(W\tilde{X_j} +b) \nonumber \\
\theta &=& \argmin L(h, X)) \nonumber 
\end{eqnarray} 

where $I$ is the embedding function. We describe it in detail as follows.

We use similar mechanisms to vectorize $U$, $T$ and $D$. For URL $U$, we break it into fixed number $K$ tokens, e.g., $20$ tokens. (The domain is a separate token. If there are less than $K$ tokens, we pad them with some null values.) Each token is then hashed into a fixed large space $M$, e.g., $10^5$. We maintain a dictionary of size $M$, in which each value is an embedding vector of fixed dimension $L$. For each selected token in $U$, we extract its corresponding embedding value from the dictionary and we pad $K$ such vectors into a final vector to represent $U$. So URL $U$ is embedded into a vector of dimension $K\cdot L$. We apply the same mechanism to $T$ and $D$ with different embedding dictionaries. The final embedded vector is of dimension $3\cdot K\cdot L$.
Some characteristic of this embedding mechanism:

\begin{itemize}
\item {\bf Unseen events.} Since our embedding is based on tokens, we can naturally handle unseen events, e.g., with new URLs. (Some websites use GUID in the URLs to make every visit with unique URL.)
\item {\bf Same tokens.} The same token appears in URL, title and description might be embedded into different vectors.
\item {\bf Cross-site correlation.} We are able to understand the similarity between different websites from the tokens in the URLs, titles and descriptions. 
\end{itemize}

The hidden representation, $h$, is mapped from the embedding through the network with the activation function, $f (\cdot{})$ as ReLu~\cite{nair2010rectified}. The weight matrix $W$ and bias $b$ are shared across the input pairs. 

The contrastive loss function, $L (\cdot, \cdot)$ is defined as
\begin{eqnarray}
L(h, X) \\
&=&  \frac{1}{2}\cdot y \cdot D(h_i, h_j) \nonumber \\
&+& \frac{1}{2} \cdot (1-y) \cdot \max\{0, m-D(h_i, h_j)\} \nonumber 
\end{eqnarray}  

$D(h_i, h_j)$ is the distance between the embedding pair $\{h_i, h_j\}$. Distance function $D$ can be customized to fulfill specific tasks. Here we use $L2$ distance. $m$ is a margin value which is greater than 0. Having a margin allows us to discard the loss from pairs are too far away and focus on the negative sample pairs that are marginally close in the current latent space.

$y$ is the label to indicate if the input pair is a positive sample, i.e., visited consecutively in one user session, or a negative one, i.e., sampled randomly from all events. Notice our loss function will enforce the embedding to pull consecutive events together, in the same time to push negative sampled events pairs to be further apart in the latent space.

%Yajun: this is confusing on the weight sharing. Same token from different data sources do not share weights.
%Considering we have heterogeneous data sources $U$, $T$ and $D$, the weight sharing layer enables the networks to learn the cross-source correlations among different domains which share the similar title/descriptions and the same word set that occurs among different URLs. For example, if two URLs are not clicked by the user consecutively but they share the similar description, then the network is still able to learn such coherence by linking these two events based on the similar word 'anchor'.     

The hidden representation layer $h$ and the embedding layer $I$ both learn the semantic representations of the user events. Another benefits of our Siamese network structure is, whenever we need the embedding, we can easily plug the whole networks as a module into the big neural network framework. To keep the contribution clear and facilitate the evaluation, we use outputs of the the embedding layer $I$ as the event vectors. 

The Siamese network structure is illustrated in Figure \ref{fig:Siamese}. We use two hidden layers instead of one in practice to leverage the different data sources, which give us lower loss during training.

Another option is to train the event embedding with word2vec e.g., Skip-gram model~\cite{mikolov2013distributed}. However, we do not see meaningful improvement over the Siamese networks. We plan to further investigate in word2vec based embedding approaches.

\subsection{Qualitative Evaluation}
% we compare two different embedding experiment settings: 
% \begin{itemize}
% \item Co-training one Siamese networks with the event triplets $E(U, T, D)$ (as shown in Figure \ref{fig:Siamese}).
% \item Training three Siamese network with the events $U, T, D$ respectively.
% \end{itemize}

It is a very challenging task to measure the quality of our event embedding scheme. We qualitatively inspect the embedding results as follows. 
As the user behavior data provide no category information, we manually selected several pivots (or say, centroid). For each pivot, we retrieve the relevant embedding representations and find the top-k nearest neighbors in the embedding space. This is motivated by the assumption that a useful representation would cluster
the domain/tokens with similar semantic contents. We generate a
subset that contains the top 80 events for selected pivots. We applied t-SNE \cite{maaten2008visualizing} with a $L2$ norm kernel to reduce the dimensionality of the item vectors to 2. Then, we color each pivots and their neighbors. 
%See Figure~\ref{fig:embedding_graph} for the outcome of the embedding.

\begin{figure}[t]
	\centering
	\includegraphics[width=0.5\textwidth]{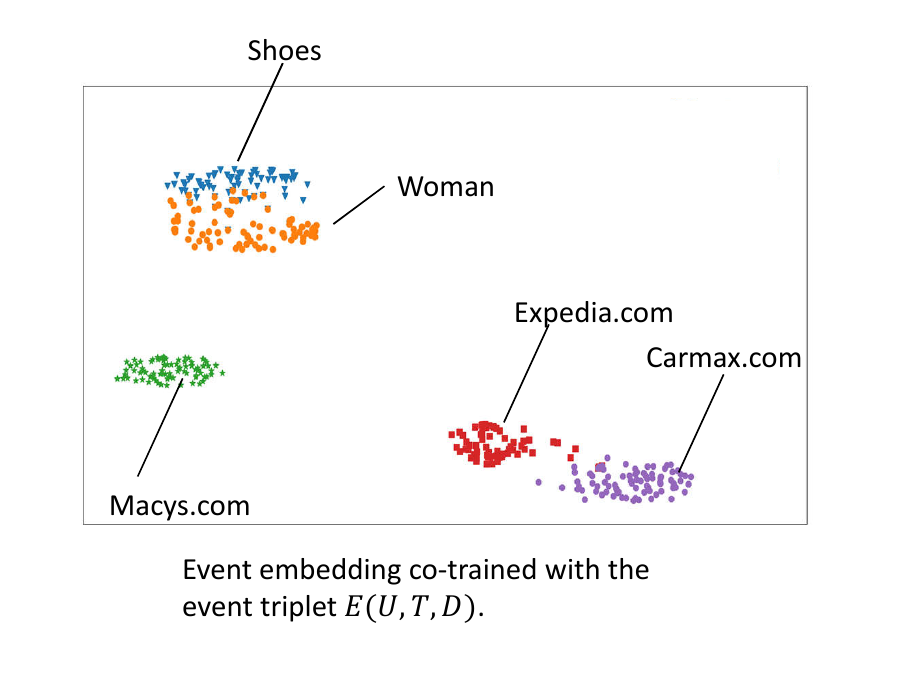}
	\caption {Visualization of embeddings of the user events.}
	\label{fig:embedding_graph}
\end{figure}

% As shown in Figure \ref{fig:embedding_graph}, the right graph illustrates the embedding trained with individual Siamese networks on each $U, T, D$. The cluster of pivots 'shoes', 'woman', and 'Expedia.com' are not significantly distinguishable in the latent space. Although the projection of the individual embeddings are naturally uncorrelated, they are still able to confuse the learning by these similar projections on different entities which represent a variety of user intention. Moreover, the coherence within each cluster is not strong as the points are loosely scatter around multiple pivots.
% * <yalding@gmail.com> 2018-02-11T23:27:26.632Z:
% 
% > The cluster of pivots 'shoes', 'woman', and 'Expedia.com' are not significantly distinguishable in the latent space
% This is not a fair comparison as the latent spaces in three embeddings have different meanings. I do not see any value for presenting two schemes if we are not able to find meaningful comparisons.
% 
% ^.
%Training the embedding with the user event triplets, we found the embedding quality is improved on two manifolds. 

As shown in the left graph in Figure \ref{fig:embedding_graph}, every selected pivots and their associated clusters are clear and well defined. 'Shoes' and 'Woman' are close to each other with some overlap, which is reasonable as the shoes are a quite big category in female consumer goods. 'expedia.com' and 'carmax.com' are also near with each other. Moreover, the nearest neighbors trained have strong coherence. We pulled the neighbors beyond the top 100 in each embeddings, the neighbors are still showing reasonable correlation to the pivots. 
%but as for the individual embeddings, much of noise and non-relevant items appears at the same distance level.  
\begin{table*}[t]
	\centering
	\caption{Demo of the event embedding results. Given a pivot domain or words, we extract their nearest neighbor}
	\begin{tabular}{c|l}
		\multicolumn{1}{l}{Pivot} &  Nearest Neighbors \\
		\midrule
		\multirow{2}[0]{*}{www.carmax.com} & www.edmunds.com, www.toyota.com, www.subaru.com, www.carfax.com,  \\
		& www.mbusa.com, automobiles.honda.com, www.chevrolet.com \\
		\midrule
		\multirow{2}[0]{*}{www.macys.com} & www.ebates.com, www.jcpenney.com, www.sephora.com, www.target.com,www.ulta.com, \\
		&  www.potterybarn.com, www.gap.com, www.urbanoutfitters.com, www.dsw.com, www.overstock.com \\
		\midrule
		\multirow{2}[0]{*}{www.expedia.com} & www.kayak.com, www.priceline.com, www.orbitz.com, www.hotels.com, www.starwoodhotels.com \\
		& www.travelocity.com, usa, www.delta.com, www.airbnb.com  \\
		\midrule
		\midrule
		\multirow{2}[0]{*}{Woman} & petite, lace, kids, gifts, baby, christmas, dresses, hair, plus, beauty, shoes,  \\
		& pants, boots, makeup, outerwear, girls, checkout, boys, sale, shoe \\
		\midrule
		\multirow{2}[0]{*}{Shoes} & boots, women, womens, apparel, mens, dress, clothing, shirt, kids, jeans,  \\
		&  jackets, beauty, shirts, wool, sets, face, checkout, jacket, girls, black \\
		\bottomrule
	\end{tabular}%
	\label{tab:emb_demo}%
\end{table*}%

In Table \ref{tab:emb_demo}, we show the examples of the selected pivots and their top nearest neighbors. The URL domains and words are clustered respectively although we trained them by projecting onto a common hidden layer together. The embeddings are learning the semantics that encodes the user browsing interest and intentions well, like Glove \cite{pennington2014glove} or DSSM \cite{huang2013learning}. Especially, the embeddings learns the specific patterns in user browsing history, like for the centroid 'woman', 'hair', 'beauty' and even 'sale' are among the closest neighbors as they also appear often when the user click and browse.  
Note that our method and scenario are correlated but also different with DSSM.
% * <yalding@gmail.com> 2018-02-11T18:11:40.272Z:
% 
% We probably do not want to mention DSSM as we are not discussing it in the related work.
%  
% ^.
% I have added the related work.

\begin{itemize}
\item DSSM trains the embedding using query-document pairs. It contains strong supervision from the search query and clicks. In native ads, as we have no related query, our methods trained the embedding that is weakly supervised by the pure user browsing history, which is more realistic and adequate in native advertising.  
\item DSSM is able to represent short sentences as the query-answer pair always appears in such continuous word sets. In native ads, either the URLs, title or short description in the user browsing history has meaningful syntactic structures. Thus we proposed the Siamese network structure instead of Skip-gram. Our method is able to embed both the words and URL domains.  
\end{itemize}

\section{Click Through Rate Prediction}
In this section, we apply our embedding results to the CTR prediction problem for native ads. In particular, for each request $\{ E_u^{-t}, a_t\}$ where $E_u^{-t}$ is the sequence of activities for the user $u$ up to time $t$ and $a_t$ is the advertisement we displayed to user $u$. Our objective is to predict the probability that user $u$ will click on $a_t$ in this request. Figure \ref{fig:flowchart} outlines the information flow.

We describe several variations to model
user activities from the browsing history of the user as the input to the CTR prediction tasks. First, we
formulate our problem and a simple word-based baseline method
and discuss the issues that they have. We then describe our 
methods of using latent embedding of user events, as was explained
in the previous section.

\begin{figure*}[t]
	\centering
	\includegraphics[width=0.8\textwidth]{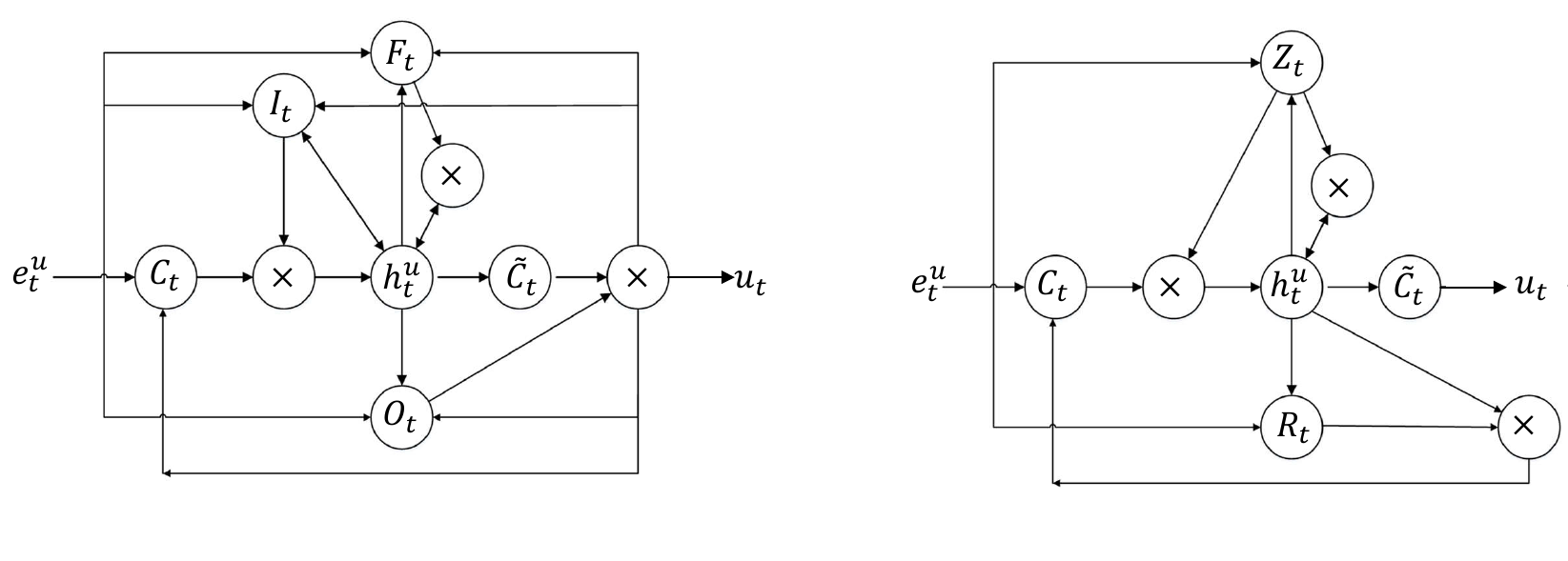}
	\caption {LSTM unit (left) and GRU unit (right).}
	\label{fig:recurrent}
\end{figure*}

\subsection{Problem Formulation}
Let $E_u$ be the entire sequence of the events for user $u$ and $E_u^{-t}$ be the sequence of events of user $u$ up to time $t$. We are given a set of triplets $\{<u_t, a_t, y_t>\}$, that user $u_t$ is served with advertisement $a_t$ at time $t$. $y_t$ indicates that the $u_t$ clicked on $a_t$ (if $1$) or not (if $0$). Our objective is:

\begin{figure}[h]
	\centering
	\includegraphics[width=0.5\textwidth]{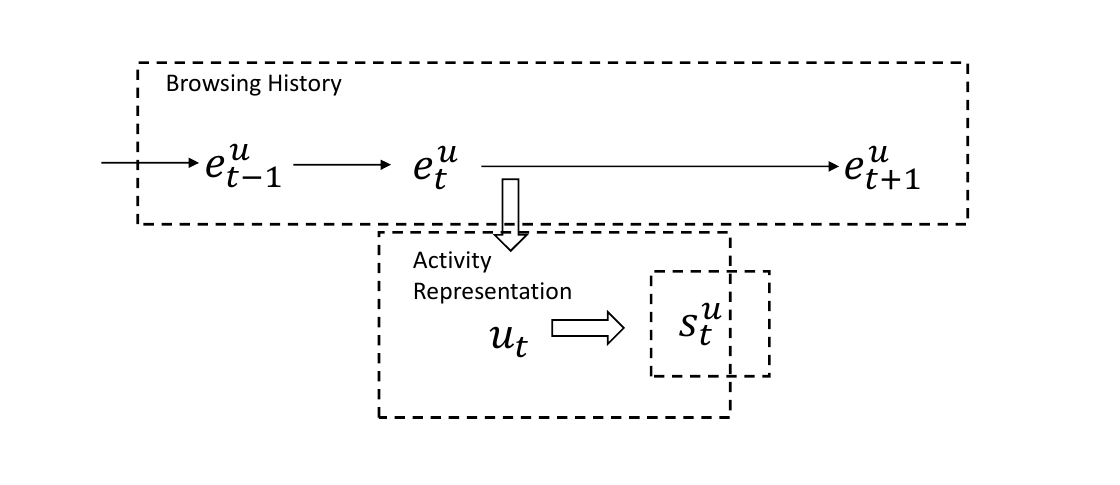}
	\caption {Flowchart of browsing history and clicks.}
	\label{fig:flowchart}
\end{figure}

%Here, $u_t$ is user state summarized from the browsing activity, which depends on user browsing history $e^u_1, e^u_2, \cdots, e^u_n $. $u_t$ should be able to represents the preference of $u$ immediately after the event $e^u_t$. As for CTR prediction in native ads, it is also important to understand and model the user's intention from the series of the browsing behavior. Our goal is to learn the user state function $F(e^u_t)$ that satisfies

\begin{eqnarray}
%u_t = F(e^u_1, e^u_2, \cdots, e^u_n) \nonumber \\
\argmin \sum_{t} L_{\theta}( y_t, \tilde{y}_t; E^{-t}_{u_t}, a_t)
\end{eqnarray} 

Where $L$ is the loss of predicted click $\tilde{y}$ and the real click $y$ given user browsing events $E_{u_t}^{-t}$ up to time $t$. $\theta$ is the model parameters. Based on this formulation, the most important part of the model is to represent the event history $E^{-t}_u$ for user $u$. In the following sections, we describe a few different choices to model the user event history.

%yajun: need a more refined way to say streaming apps.
%Note that When considering the constrained response time of our near real-time native ads streaming system, the function $F$ must be scalable to large amount of streaming data. Because browsing events are streaming fast, it is impossible to catch all the events from the users ${e^u_t, u\in U, t\in T }$. Thus, it is necessary to update the state $u_t$ for millions of users in near real-time until the recommended list of ads from visiting our service page is displayed. When design the baseline and our proposing model, we follow the principal to design the model that is suitable for near real-time streaming systems.

\subsection{Single Event Representation}
\label{ss:singleeventembedding}
We explore three different event representations.
\begin{enumerate}
	\item  {\bf Bag of Word (BoW).} The bag-of-word based model introduced in \cite{okura2017embedding} are proposed in our experiment as the baseline.  An event $e=\{U, T, D\}$ is represented by a collection of words tokenized from the triplets. The user state is learned from a collection of words included in the last event (last-1 event baseline) before time $t$ or a fixed number of events  (last-N event) of the user before time $t$. We use a feed forward networks to learn a non-linear mapping on the collection of the words to represent the user history.
	\item {\bf Co-trained Event Embeddings (Embedding).} We use the network structure in Section~\ref{sec:event_vector} to embed the events. However, the weights of the embedding layers is co-trained together with the overall network for our CTR task.
    \item {\bf Pretrained Embedding.} The embedding vector is pretrained by the Siamese networks in Section~\ref{sec:event_vector}.
\end{enumerate}

BoW is simple and fast to calculate, but suffers from the sparseness of the representation. The co-trained event embedding has a large number of weights and hence the model complexity is very high. In our CTR prediction task, the labeled data is much smaller than the users' browsing history, which makes the co-training of the embedding vector a lot challenging. On the other hand, the pretrained event embedding is trained in self-supervised manner with sufficient labeled data. As we can see later, pretrained embedding has more generalization power.

\begin{figure*}[t]
	\centering
	\includegraphics[width=0.9\textwidth]{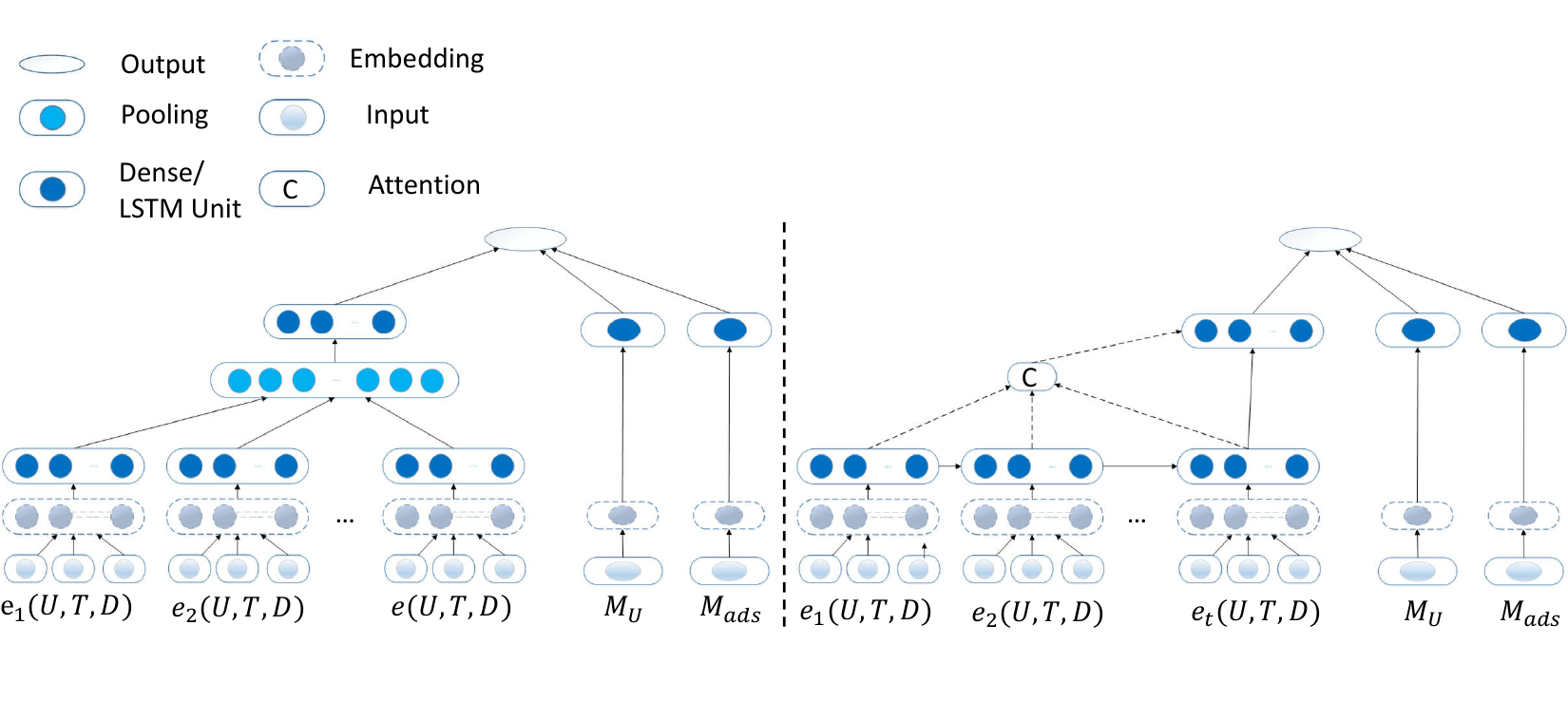}
	\caption {Network structure of the bag-of-events and the recurrent (attentional) model.}
	\label{fig:CTR_net}
\end{figure*}

Now we describe how to represent users' history based on the single event representations.

\subsection{Bag-of-Events Model (BoE)}
Following the popular model structure introduced in ~\cite{shan2016deep,covington2016deep,zhou2017deep}, our base bag-of-events model has two major steps:

\begin{enumerate}
	\item Use one of the single event embedding approach in Section~\ref{ss:singleeventembedding}.
	\item Apply a feed forward networks with ReLU to fit the click response. Notice that the input contains multiple user events, we add an averaging pooling layer to summarize the sequence and get a fixed size vector.
\end{enumerate}

We use BoE$(K)$ to represent the bag of events model with last $K$ events included before the pooling layer.

%As illustrated in the left graph of Figure \ref{fig:CTR_net}, the base model works reasonably well in many of the practical systems and equipped with simpler model complexity, which serves as the main baseline in our native advertising systems.

A decay model is proposed in~\cite{okura2017embedding} as a baseline where  $\alpha$ is a parameter vector that represents the strength
of time decay. However, the decay model is not showing any gain in their experiments. Since our (attentional) recurrent model is focusing on learning the temporal dependencies, we skip this variation in our experiments.  

\subsection{Recurrent Neural Network Models}
Given that the user's history consists of an event sequence, it is natural to consider recurrent neural networks (RNN) to model the user. More specifically, in such RNN networks, the user state $s_u^t$ is determined by the event $e^u_t$ and previous state $s_u^{t-1}$ as:
\begin{eqnarray}
s_u^t = f(e_u^t, s_u^{t-1}). \nonumber 
\end{eqnarray} 

%Naturally it is the form of the simple recurrent model which is formulated by
%\begin{eqnarray}
%u_t = \phi (W^{in}e^u_t, W^{out}u_{t-1} + b) \nonumber 
%\end{eqnarray} 
%where $\phi(\cdot)$ is the activation function of hyperbolic tangent $tanh(\cdot)$. Weight matrices $W$ in and bias term $b$ are trained using back-prorogation through time and initial state vector $u_0$ is a common initial vector that does not depend on $u$.

In order to learn long temporal dependencies and avoid gradient vanishing and explosion problems, we employ two popular RNN networks: the long short-term memory networks (LSTM)~\cite{hochreiter1997long} and the gated recurrent networks~\cite{cho2014properties}.

\subsubsection{Long short-term Memory Network}
The LSTM unit is formulated as
\begin{eqnarray}
I_t = \phi (W^{in}_{I}e^u_t + W^{out}_{I}u_{t-1} + W^V_Ih^u_{t-1}+b_I) \nonumber \\
F_t = \phi (W^{in}_{F}e^u_t + W^{out}_{F}u_{t-1} + W^V_Fh^u_{t-1}+b_F) \nonumber \\
O_t = \phi (W^{in}_{O}e^u_t + W^{out}_{O}u_{t-1} + W^V_Oh^u_{t-1}+b_O) \nonumber \\
C_t = \tau (W^{in}_Ce^u_t +W^{out}_Cu_{t-1}+b_C) \nonumber \\
h^u_t = I_t\odot C_t + F_t \odot h^u_{t-1}  \\
\tilde{C_t} = \tau (W^V_{\tilde{C}}h^u_t +b_{\tilde{C}})  \\
u_t = O_t \odot \tilde{C_t} 
\end{eqnarray} 

where $\phi (\cdot)$ is element-wise sigmoid function. $h^u_t$ is the memory state. The information flows from the input of the user browsing browsing activity  to output (user state). The input $e^u_t$ is encoded using pretrained event embedding vectors in section \ref{sec:event_vector} or can be randomly initialized and trained with the CTR networks. The event vector flow merged into the previous hidden and decoded to the event vector space as the user state.

The browsing information flow is controlled by three gates, input gate $I_t$, forget gate $F_t$, and output gate $O_t$. The idea is that the input gate filters unnecessary input browsing activity to construct a user state, for example the concept drift due to the sudden interest change. The forget gate represents the decline in interest by the user, which is actually a more adaptive time decay effect than the exponential discount factor. The output gate filters the contents that should not be focused on in the next session.

\subsubsection{Gated Recurrent Network}
The gated recurrent unit (GRU) was proposed in ~\cite{cho2014properties} to make each recurrent unit adaptively capture the dependencies of different time scales, which is another effective learning structure to avoid gradient vanishing and explosion. The GRU unit is formulated as:
\begin{eqnarray}
Z_t = \phi (W^{in}_{Z}e^u_t + W^{V}_{Z}h_{t-1} + b_Z) \nonumber \\
R_t = \phi (W^{in}_{R}e^u_t + W^{V}_{R}h_{t-1} + b_R) \nonumber \\
C_t = \tau (W^{in}_Ce^u_t +W^{out}_C(R_t\odot h^{t-1})+b_C) \nonumber \\
h^u_t = Z_t\odot C_t + (1-Z_t) \odot h^u_{t-1}  \\
\tilde{C_t} = \tau (W^V_{\tilde{C}}h^u_t +b_{\tilde{C}})  \\
u_t = \tilde{C_t} 
\end{eqnarray} 

The important difference is that the gate $Z_t$ in this model plays
the role of two gates, i.e., input and forget gate
in the LSTM-based model, which happens to make the upper bound of the hidden state $h$ as $\max(||h^u_0||_\infty, sup_x|\tau(x)|)$. Thus, for any length of the long input sequences, the upper bound tends to be limited by a constant. In practice, LSTM model is always trained with gradient clipping ~\cite{pascanu2013difficulty} where GRU unit is always trained without clipping. Figure \ref{fig:recurrent} shows the structure of LSTM unit and GRU unit.

Note that except for one original GRU layer, we add another fully
connected layer as suggested by~\cite{okura2017embedding}, which makes the network easy to adopt the proposed attention model.

\subsubsection{Attention Network}
Attention model is proposed in ~\cite{bahdanau2014neural,graves2013generating} in machine translation and sequence learning tasks, and later successfully applied to recommendation systems ~\cite{Wangy2017Attentionbased}.  

We build a context vector $c_i$ that depends on a sequence of annotations $(h_1, \cdots, h_t)$ to which a recurrent encoder maps the input browsing events. Each $h_i$ contains information about the whole input sequence with a strong focus on the parts surrounding the $i$-th event in the input sequence. To better capture the different contribution scale of various contextual
events, the attention vector learns the integration
weights on each $E_i$ instead of using the last state only. 

The context vector $c_i$ is computed as the weighted sum of the state sequences $h_i$
\begin{eqnarray}
c_i = \sum_{j=1}^{t} \alpha_jh_j \nonumber \\
\end{eqnarray} 

The weight $\alpha$ of state $h_j$ is computed by
\begin{eqnarray}
\alpha_{ij} = \frac{\exp{e_{j}}}{\sum_{j=1}^t \exp(e_j)} 
\nonumber \\
e_{j} = \tau(h_j) \nonumber 
\end{eqnarray} 

Where $\tau$ is the hyper-tangent function. The context vector directly computes a soft attention, which allows the gradient of the cost function to be back-propagated through. Indeed, the context vector is a weighted sum of all the annotations as computing an expected annotation, which assign the probability that
the target event ${e^u_i}$ is assigned to the final user states. All the three models are illustrated in Figure \ref{fig:CTR_net}.

\subsection{Experiments and Results}
\subsubsection{Dataset}
To train the Siamese network for event embedding, we sampled approximately one week user browsing history and build about 600 million event pairs which has been visited consecutively from our browsing data in November, 2017. To train the CTR model, we random sampled about 100 million browsing and click entries. About 60 million is used as training data, leave the remaining 40 million as validation and test set evenly. 

For CTR prediction, we extract the last $K=10$  browsing events before the clicks/non-clicks on the advertisement to model the user intention. We also test the baseline that only uses the last event. The user meta data $M_u$, e.g., user ID and the advertisement meta data  $M_{ads}$, e.g., title, description, URL are used to predict the click. The full dataset has about $0.36\%$ events has label 1 (clicked).

\subsubsection{Evaluation Metrics}
To measure CTR prediction quality, we compute the area under receiver operator curve (AUC)~\cite{fawcett2006introduction} as the labels/clicks are highly skewed. For the clicks in test set, we calculate AUC using the predicted ranking formed by the estimated click probability output by the model and the ground-truth label, which is the presence or
absence of clicks on the native advertisement. We report final
AUC as an average and standard deviation over five runs.

\subsubsection{Experiment Setting and Result}
We fully compare different baseline settings and our proposed recurrent models with pretrained event embeddings. The embedding vector size are fixed at $128$. All the dense layer have $256$ units. We trained all the neural networks with Adam ~\cite{kingma2014adam} with the
initial learning rate was $0.0001$ and mini-batch is $128$. The dimensions of internal state $h^u_t$ is $256$ in both LSTM and GRU. Dropout ratio is $0.5$ at the last layer in all the neural networks models to alleviate overfitting. These parameters are determined using the validation dataset.
We used the gradient clipping technique with bound $5$ to avoid gradient explosion in LSTM. GRU did not cause gradient explosion when this technique was not used in these experiments. 

\begin{table}[h]
  \centering
  \caption{Results of the click prediction. Values indicate average of metrics and $99\%$ confidence intervals in five runs.}
    \begin{tabular}{l|l}
    \toprule
    Model & AUC  \\
    \toprule
    BoW - BoE(1) & 0.623 $\pm$ 0.004 \\
    BoW - BoE(10) & 0.631 $\pm$ 0.005 \\
    Pretrained embedding  - BoE(1) & 0.630 $\pm$ 0.005 \\
    Pretrained embedding  - BoE(10) & 0.642 $\pm$ 0.006 \\
    \midrule
    \midrule
    Embedding - LSTM  & 0.679 $\pm$ 0.005 \\
    Embedding - GRU & 0.683 $\pm$ 0.004 \\
    Pretrained embedding  - LSTM & 0.695 $\pm$ 0.003 \\
    Pretrained embedding  - GRU & 0.697 $\pm$ 0.002 \\
    Pretrained embedding  - LSTM - attention & \textbf{0.699} $\pm$ \textbf{0.005} \\
    Pretrained embedding  - GRU - attention & \textbf{0.708} $\pm$ \textbf{0.004} \\
    \bottomrule
    \end{tabular}%
  \label{tab:CTR_result}%
\end{table}%

Table \ref{tab:CTR_result} lists all the AUC metrics with different experiment settings. BoE is short the Bag-of-Events model. The brackets $(1)$ indicates that we use the only last user event to model the user activities. 'Embedding' means we randomly initialize and train the embedding within the click prediction training procedure. We also test the case to use the pretrained embedding to represent the user events.  

For the Bag of Events models, the model with last 10 events improves over the single event case. Also by employing the pretrained embedding, we are able to improve the AUC slightly further.

In the second half of Table~\ref{tab:CTR_result}, RNN based models, e.g., LSTM and GRU achieve significantly better results than Bag-of-Events models. This indicates that the RNN networks generally can better summarize sequential user activities. 
We also observe that the pretrained embedding is able to improve the AUC for both LSTM and GRU models. The attention layer can further improve the performance. We notice that GRU models consistently outperform LSTM networks in all our settings.

\begin{table}[h]
  \centering
  \caption{AUC score of the recurrent model with different user event history length.}
    \begin{tabular}{l|l}
    \toprule
    Model & AUC  \\
    \toprule
    Pretrained embedding  - LSTM - attention (5) & 0.695 $\pm$ 0.002 \\
    Pretrained embedding  - LSTM - attention (10) & 0.699 $\pm$ 0.003 \\
    Pretrained embedding  - LSTM - attention (20) & 0.698 $\pm$ 0.003 \\
    \midrule
    Pretrained embedding  - GRU - attention (5) & 0.702 $\pm$ 0.004 \\
    Pretrained embedding  - GRU - attention (10) & 0.711 $\pm$ 0.004 \\
    Pretrained embedding  - GRU - attention (20) & 0.711 $\pm$ 0.003 \\
    \midrule
    \end{tabular}%
  \label{tab:length}%
\end{table}%

In Table \ref{tab:length}, we explore how the length of the user activities would impact the performance of the models. We see if the length is short, the performance is slightly worse due to the insufficient learning of the user states. However, the models hardly benefit from the long user history beyond $10$ events. We plan to investigate further how to model long term history more effectively.

One interesting observation we have is the generalization power of the pretrained event embedding model. From Table~\ref{tab:CTR_delta}, while training AUC decreases with pretrained embedding vectors, the testing AUC actually increases. This matches our intuition as we could have overfitting issue when co-training the event vectors.

\begin{table}[h]
  \centering
  \caption{The difference of the mean AUC score between training and test with different models.}
    \begin{tabular}{l|l|l}
    \toprule
    Model and AUC & Training & Test  \\
    \toprule
    Embedding - GRU  & 0.746  &	0.683 \\
    Pretrained embedding  - GRU & 0.735	& 0.697 \\
    Pretrained embedding  - GRU - attention &  0.741 & 0.711 \\
    \bottomrule
    \end{tabular}%
  \label{tab:CTR_delta}%
\end{table}%

\section{Conclusion}
In this paper, we propose a large-scale event embedding scheme to encode the user browsing history by training a Siamese network with weak supervision on the user consecutive events. With the pretrained embedding, we explore recurrent neural network models to model the user history, thus to improve the CTR prediction in native advertising. 

We found that the our embedding scheme is able to learn reasonable semantics of both domains and words by training from the URL, title and description of the user events. 

We also show the improvements in the native advertising CTR prediction task by using recurrent neural network models with the pretrained embedding and attention layer to model the user history. Our experiments showed that the event embedding and our designed recurrent attentional models significantly outperform the baseline and other variants.

%Furthermore, future work can improve multiple facets of our model. For instance, the attention vector is addressed on the input sequences, which is constrained by the user event length. One possible solution is to learn a external memory that allows multi-probe read and write to enable the multi-intention learning beyond any browsing sequences. 

\bibliographystyle{ACM-Reference-Format}
\bibliography{sigproc}


\begin{thebibliography}{32}

%%% the Web address.

\ifx \showCODEN    \undefined \def \showCODEN     #1{\unskip}     \fi
\ifx \showDOI      \undefined \def \showDOI       #1{#1}\fi
\ifx \showISBNx    \undefined \def \showISBNx     #1{\unskip}     \fi
\ifx \showISBNxiii \undefined \def \showISBNxiii  #1{\unskip}     \fi
\ifx \showISSN     \undefined \def \showISSN      #1{\unskip}     \fi
\ifx \showLCCN     \undefined \def \showLCCN      #1{\unskip}     \fi
\ifx \shownote     \undefined \def \shownote      #1{#1}          \fi
\ifx \showarticletitle \undefined \def \showarticletitle #1{#1}   \fi
\ifx \showURL      \undefined \def \showURL       {\relax}        \fi
% invisible to TeX
\providecommand\bibfield[2]{#2}
\providecommand\bibinfo[2]{#2}
\providecommand\natexlab[1]{#1}
\providecommand\showeprint[2][]{arXiv:#2}

\bibitem[\protect\citeauthoryear{??}{bin}{[n. d.]}]%
        {bingintent}
 \bibinfo{year}{[n. d.]}\natexlab{}.
\newblock \bibinfo{title}{Bing Intent Ads}.
\newblock
  \bibinfo{howpublished}{\url{https://advertise.bingads.microsoft.com/en-us/solutions/ad-products/bing-intent-ads}}.
    (\bibinfo{year}{[n. d.]}).
\newblock


\bibitem[\protect\citeauthoryear{??}{fac}{[n. d.]}]%
        {facebookads}
 \bibinfo{year}{[n. d.]}\natexlab{}.
\newblock \bibinfo{title}{Facebook Ads}.
\newblock
  \bibinfo{howpublished}{\url{https://www.facebook.com/business/products/ads}}.
    (\bibinfo{year}{[n. d.]}).
\newblock


\bibitem[\protect\citeauthoryear{??}{sha}{[n. d.]}]%
        {sharethrough}
 \bibinfo{year}{[n. d.]}\natexlab{}.
\newblock \bibinfo{title}{NATIVE ADS VS BANNER ADS}.
\newblock
  \bibinfo{howpublished}{\url{https://www.sharethrough.com/resources/in-feed-ads-vs-banner-ads/}}.
    (\bibinfo{year}{[n. d.]}).
\newblock


\bibitem[\protect\citeauthoryear{??}{yah}{[n. d.]}]%
        {yahoogemini}
 \bibinfo{year}{[n. d.]}\natexlab{}.
\newblock \bibinfo{title}{Yahoo Gemini Ads}.
\newblock
  \bibinfo{howpublished}{\url{https://gemini.yahoo.com/advertiser/home}}.
  (\bibinfo{year}{[n. d.]}).
\newblock


\bibitem[\protect\citeauthoryear{Bahdanau, Cho, and Bengio}{Bahdanau
  et~al\mbox{.}}{2014}]%
        {bahdanau2014neural}
\bibfield{author}{\bibinfo{person}{Dzmitry Bahdanau},
  \bibinfo{person}{Kyunghyun Cho}, {and} \bibinfo{person}{Yoshua Bengio}.}
  \bibinfo{year}{2014}\natexlab{}.
\newblock \showarticletitle{Neural machine translation by jointly learning to
  align and translate}.
\newblock \bibinfo{journal}{\emph{arXiv preprint arXiv:1409.0473}}
  (\bibinfo{year}{2014}).
\newblock


\bibitem[\protect\citeauthoryear{Barkan and Koenigstein}{Barkan and
  Koenigstein}{2016}]%
        {barkan2016item2vec}
\bibfield{author}{\bibinfo{person}{Oren Barkan} {and} \bibinfo{person}{Noam
  Koenigstein}.} \bibinfo{year}{2016}\natexlab{}.
\newblock \showarticletitle{Item2vec: neural item embedding for collaborative
  filtering}. In \bibinfo{booktitle}{\emph{Machine Learning for Signal
  Processing (MLSP), 2016 IEEE 26th International Workshop on}}. IEEE,
  \bibinfo{pages}{1--6}.
\newblock


\bibitem[\protect\citeauthoryear{Cheng, Koc, Harmsen, Shaked, Chandra, Aradhye,
  Anderson, Corrado, Chai, Ispir, et~al\mbox{.}}{Cheng et~al\mbox{.}}{2016}]%
        {cheng2016wide}
\bibfield{author}{\bibinfo{person}{Heng-Tze Cheng}, \bibinfo{person}{Levent
  Koc}, \bibinfo{person}{Jeremiah Harmsen}, \bibinfo{person}{Tal Shaked},
  \bibinfo{person}{Tushar Chandra}, \bibinfo{person}{Hrishi Aradhye},
  \bibinfo{person}{Glen Anderson}, \bibinfo{person}{Greg Corrado},
  \bibinfo{person}{Wei Chai}, \bibinfo{person}{Mustafa Ispir}, {et~al\mbox{.}}}
  \bibinfo{year}{2016}\natexlab{}.
\newblock \showarticletitle{Wide \& deep learning for recommender systems}. In
  \bibinfo{booktitle}{\emph{Proceedings of the 1st Workshop on Deep Learning
  for Recommender Systems}}. ACM, \bibinfo{pages}{7--10}.
\newblock


\bibitem[\protect\citeauthoryear{Cho, Van~Merri{\"e}nboer, Bahdanau, and
  Bengio}{Cho et~al\mbox{.}}{2014}]%
        {cho2014properties}
\bibfield{author}{\bibinfo{person}{Kyunghyun Cho}, \bibinfo{person}{Bart
  Van~Merri{\"e}nboer}, \bibinfo{person}{Dzmitry Bahdanau}, {and}
  \bibinfo{person}{Yoshua Bengio}.} \bibinfo{year}{2014}\natexlab{}.
\newblock \showarticletitle{On the properties of neural machine translation:
  Encoder-decoder approaches}.
\newblock \bibinfo{journal}{\emph{arXiv preprint arXiv:1409.1259}}
  (\bibinfo{year}{2014}).
\newblock


\bibitem[\protect\citeauthoryear{Chopra, Hadsell, and LeCun}{Chopra
  et~al\mbox{.}}{2005}]%
        {chopra2005learning}
\bibfield{author}{\bibinfo{person}{Sumit Chopra}, \bibinfo{person}{Raia
  Hadsell}, {and} \bibinfo{person}{Yann LeCun}.}
  \bibinfo{year}{2005}\natexlab{}.
\newblock \showarticletitle{Learning a similarity metric discriminatively, with
  application to face verification}. In \bibinfo{booktitle}{\emph{Computer
  Vision and Pattern Recognition, 2005. CVPR 2005. IEEE Computer Society
  Conference on}}, Vol.~\bibinfo{volume}{1}. IEEE, \bibinfo{pages}{539--546}.
\newblock


\bibitem[\protect\citeauthoryear{Covington, Adams, and Sargin}{Covington
  et~al\mbox{.}}{2016}]%
        {covington2016deep}
\bibfield{author}{\bibinfo{person}{Paul Covington}, \bibinfo{person}{Jay
  Adams}, {and} \bibinfo{person}{Emre Sargin}.}
  \bibinfo{year}{2016}\natexlab{}.
\newblock \showarticletitle{Deep neural networks for youtube recommendations}.
  In \bibinfo{booktitle}{\emph{Proceedings of the 10th ACM Conference on
  Recommender Systems}}. ACM, \bibinfo{pages}{191--198}.
\newblock


\bibitem[\protect\citeauthoryear{Fawcett}{Fawcett}{2006}]%
        {fawcett2006introduction}
\bibfield{author}{\bibinfo{person}{Tom Fawcett}.}
  \bibinfo{year}{2006}\natexlab{}.
\newblock \showarticletitle{An introduction to ROC analysis}.
\newblock \bibinfo{journal}{\emph{Pattern recognition letters}}
  \bibinfo{volume}{27}, \bibinfo{number}{8} (\bibinfo{year}{2006}),
  \bibinfo{pages}{861--874}.
\newblock


\bibitem[\protect\citeauthoryear{Gai, Zhu, Li, Liu, and Wang}{Gai
  et~al\mbox{.}}{2017}]%
        {gai2017learning}
\bibfield{author}{\bibinfo{person}{Kun Gai}, \bibinfo{person}{Xiaoqiang Zhu},
  \bibinfo{person}{Han Li}, \bibinfo{person}{Kai Liu}, {and}
  \bibinfo{person}{Zhe Wang}.} \bibinfo{year}{2017}\natexlab{}.
\newblock \showarticletitle{Learning Piece-wise Linear Models from Large Scale
  Data for Ad Click Prediction}.
\newblock \bibinfo{journal}{\emph{arXiv preprint arXiv:1704.05194}}
  (\bibinfo{year}{2017}).
\newblock


\bibitem[\protect\citeauthoryear{Graves}{Graves}{2013}]%
        {graves2013generating}
\bibfield{author}{\bibinfo{person}{Alex Graves}.}
  \bibinfo{year}{2013}\natexlab{}.
\newblock \showarticletitle{Generating sequences with recurrent neural
  networks}.
\newblock \bibinfo{journal}{\emph{arXiv preprint arXiv:1308.0850}}
  (\bibinfo{year}{2013}).
\newblock


\bibitem[\protect\citeauthoryear{He, Liao, Zhang, Nie, Hu, and Chua}{He
  et~al\mbox{.}}{2017}]%
        {he2017neural}
\bibfield{author}{\bibinfo{person}{Xiangnan He}, \bibinfo{person}{Lizi Liao},
  \bibinfo{person}{Hanwang Zhang}, \bibinfo{person}{Liqiang Nie},
  \bibinfo{person}{Xia Hu}, {and} \bibinfo{person}{Tat-Seng Chua}.}
  \bibinfo{year}{2017}\natexlab{}.
\newblock \showarticletitle{Neural collaborative filtering}. In
  \bibinfo{booktitle}{\emph{Proceedings of the 26th International Conference on
  World Wide Web}}. International World Wide Web Conferences Steering
  Committee, \bibinfo{pages}{173--182}.
\newblock


\bibitem[\protect\citeauthoryear{Hochreiter and Schmidhuber}{Hochreiter and
  Schmidhuber}{1997}]%
        {hochreiter1997long}
\bibfield{author}{\bibinfo{person}{Sepp Hochreiter} {and}
  \bibinfo{person}{J{\"u}rgen Schmidhuber}.} \bibinfo{year}{1997}\natexlab{}.
\newblock \showarticletitle{Long short-term memory}.
\newblock \bibinfo{journal}{\emph{Neural computation}} \bibinfo{volume}{9},
  \bibinfo{number}{8} (\bibinfo{year}{1997}), \bibinfo{pages}{1735--1780}.
\newblock


\bibitem[\protect\citeauthoryear{Huang, He, Gao, Deng, Acero, and Heck}{Huang
  et~al\mbox{.}}{2013}]%
        {huang2013learning}
\bibfield{author}{\bibinfo{person}{Po-Sen Huang}, \bibinfo{person}{Xiaodong
  He}, \bibinfo{person}{Jianfeng Gao}, \bibinfo{person}{Li Deng},
  \bibinfo{person}{Alex Acero}, {and} \bibinfo{person}{Larry Heck}.}
  \bibinfo{year}{2013}\natexlab{}.
\newblock \showarticletitle{Learning deep structured semantic models for web
  search using clickthrough data}. In \bibinfo{booktitle}{\emph{Proceedings of
  the 22nd ACM international conference on Conference on information \&
  knowledge management}}. ACM, \bibinfo{pages}{2333--2338}.
\newblock


\bibitem[\protect\citeauthoryear{Kingma and Ba}{Kingma and Ba}{2014}]%
        {kingma2014adam}
\bibfield{author}{\bibinfo{person}{Diederik~P Kingma} {and}
  \bibinfo{person}{Jimmy Ba}.} \bibinfo{year}{2014}\natexlab{}.
\newblock \showarticletitle{Adam: A method for stochastic optimization}.
\newblock \bibinfo{journal}{\emph{arXiv preprint arXiv:1412.6980}}
  (\bibinfo{year}{2014}).
\newblock


\bibitem[\protect\citeauthoryear{Maaten and Hinton}{Maaten and Hinton}{2008}]%
        {maaten2008visualizing}
\bibfield{author}{\bibinfo{person}{Laurens van~der Maaten} {and}
  \bibinfo{person}{Geoffrey Hinton}.} \bibinfo{year}{2008}\natexlab{}.
\newblock \showarticletitle{Visualizing data using t-SNE}.
\newblock \bibinfo{journal}{\emph{Journal of Machine Learning Research}}
  \bibinfo{volume}{9}, \bibinfo{number}{Nov} (\bibinfo{year}{2008}),
  \bibinfo{pages}{2579--2605}.
\newblock


\bibitem[\protect\citeauthoryear{{Mikolov}, {Chen}, {Corrado}, and
  {Dean}}{{Mikolov} et~al\mbox{.}}{2013}]%
        {word2vec}
\bibfield{author}{\bibinfo{person}{T. {Mikolov}}, \bibinfo{person}{K. {Chen}},
  \bibinfo{person}{G. {Corrado}}, {and} \bibinfo{person}{J. {Dean}}.}
  \bibinfo{year}{2013}\natexlab{}.
\newblock \showarticletitle{{Efficient Estimation of Word Representations in
  Vector Space}}.
\newblock \bibinfo{journal}{\emph{ArXiv e-prints}} (\bibinfo{date}{Jan.}
  \bibinfo{year}{2013}).
\newblock
\showeprint[arxiv]{cs.CL/1301.3781}


\bibitem[\protect\citeauthoryear{Mikolov, Sutskever, Chen, Corrado, and
  Dean}{Mikolov et~al\mbox{.}}{2013}]%
        {mikolov2013distributed}
\bibfield{author}{\bibinfo{person}{Tomas Mikolov}, \bibinfo{person}{Ilya
  Sutskever}, \bibinfo{person}{Kai Chen}, \bibinfo{person}{Greg~S Corrado},
  {and} \bibinfo{person}{Jeff Dean}.} \bibinfo{year}{2013}\natexlab{}.
\newblock \showarticletitle{Distributed representations of words and phrases
  and their compositionality}. In \bibinfo{booktitle}{\emph{Advances in neural
  information processing systems}}. \bibinfo{pages}{3111--3119}.
\newblock


\bibitem[\protect\citeauthoryear{Nair and Hinton}{Nair and Hinton}{2010}]%
        {nair2010rectified}
\bibfield{author}{\bibinfo{person}{Vinod Nair} {and}
  \bibinfo{person}{Geoffrey~E Hinton}.} \bibinfo{year}{2010}\natexlab{}.
\newblock \showarticletitle{Rectified linear units improve restricted boltzmann
  machines}. In \bibinfo{booktitle}{\emph{Proceedings of the 27th international
  conference on machine learning (ICML-10)}}. \bibinfo{pages}{807--814}.
\newblock


\bibitem[\protect\citeauthoryear{Okura, Tagami, Ono, and Tajima}{Okura
  et~al\mbox{.}}{2017}]%
        {okura2017embedding}
\bibfield{author}{\bibinfo{person}{Shumpei Okura}, \bibinfo{person}{Yukihiro
  Tagami}, \bibinfo{person}{Shingo Ono}, {and} \bibinfo{person}{Akira Tajima}.}
  \bibinfo{year}{2017}\natexlab{}.
\newblock \showarticletitle{Embedding-based News Recommendation for Millions of
  Users}. In \bibinfo{booktitle}{\emph{Proceedings of the 23rd ACM SIGKDD
  International Conference on Knowledge Discovery and Data Mining}}. ACM,
  \bibinfo{pages}{1933--1942}.
\newblock


\bibitem[\protect\citeauthoryear{Pascanu, Mikolov, and Bengio}{Pascanu
  et~al\mbox{.}}{2013}]%
        {pascanu2013difficulty}
\bibfield{author}{\bibinfo{person}{Razvan Pascanu}, \bibinfo{person}{Tomas
  Mikolov}, {and} \bibinfo{person}{Yoshua Bengio}.}
  \bibinfo{year}{2013}\natexlab{}.
\newblock \showarticletitle{On the difficulty of training recurrent neural
  networks}. In \bibinfo{booktitle}{\emph{International Conference on Machine
  Learning}}. \bibinfo{pages}{1310--1318}.
\newblock


\bibitem[\protect\citeauthoryear{Pennington, Socher, and Manning}{Pennington
  et~al\mbox{.}}{2014}]%
        {pennington2014glove}
\bibfield{author}{\bibinfo{person}{Jeffrey Pennington},
  \bibinfo{person}{Richard Socher}, {and} \bibinfo{person}{Christopher
  Manning}.} \bibinfo{year}{2014}\natexlab{}.
\newblock \showarticletitle{Glove: Global vectors for word representation}. In
  \bibinfo{booktitle}{\emph{Proceedings of the 2014 conference on empirical
  methods in natural language processing (EMNLP)}}.
  \bibinfo{pages}{1532--1543}.
\newblock


\bibitem[\protect\citeauthoryear{Rendle}{Rendle}{2010}]%
        {rendle2010factorization}
\bibfield{author}{\bibinfo{person}{Steffen Rendle}.}
  \bibinfo{year}{2010}\natexlab{}.
\newblock \showarticletitle{Factorization machines}. In
  \bibinfo{booktitle}{\emph{Data Mining (ICDM), 2010 IEEE 10th International
  Conference on}}. IEEE, \bibinfo{pages}{995--1000}.
\newblock


\bibitem[\protect\citeauthoryear{Saxe and Berlin}{Saxe and Berlin}{2017}]%
        {saxe2017expose}
\bibfield{author}{\bibinfo{person}{Joshua Saxe} {and}
  \bibinfo{person}{Konstantin Berlin}.} \bibinfo{year}{2017}\natexlab{}.
\newblock \showarticletitle{eXpose: A Character-Level Convolutional Neural
  Network with Embeddings For Detecting Malicious URLs, File Paths and Registry
  Keys}.
\newblock \bibinfo{journal}{\emph{arXiv preprint arXiv:1702.08568}}
  (\bibinfo{year}{2017}).
\newblock


\bibitem[\protect\citeauthoryear{Shan, Hoens, Jiao, Wang, Yu, and Mao}{Shan
  et~al\mbox{.}}{2016}]%
        {shan2016deep}
\bibfield{author}{\bibinfo{person}{Ying Shan}, \bibinfo{person}{T~Ryan Hoens},
  \bibinfo{person}{Jian Jiao}, \bibinfo{person}{Haijing Wang},
  \bibinfo{person}{Dong Yu}, {and} \bibinfo{person}{JC Mao}.}
  \bibinfo{year}{2016}\natexlab{}.
\newblock \showarticletitle{Deep crossing: Web-scale modeling without manually
  crafted combinatorial features}. In \bibinfo{booktitle}{\emph{Proceedings of
  the 22nd ACM SIGKDD International Conference on Knowledge Discovery and Data
  Mining}}. ACM, \bibinfo{pages}{255--262}.
\newblock


\bibitem[\protect\citeauthoryear{Vincent, Larochelle, Bengio, and
  Manzagol}{Vincent et~al\mbox{.}}{2008}]%
        {Vincent:2008:ECR:1390156.1390294}
\bibfield{author}{\bibinfo{person}{Pascal Vincent}, \bibinfo{person}{Hugo
  Larochelle}, \bibinfo{person}{Yoshua Bengio}, {and}
  \bibinfo{person}{Pierre-Antoine Manzagol}.} \bibinfo{year}{2008}\natexlab{}.
\newblock \showarticletitle{Extracting and Composing Robust Features with
  Denoising Autoencoders}. In \bibinfo{booktitle}{\emph{Proceedings of the 25th
  International Conference on Machine Learning}} \emph{(\bibinfo{series}{ICML
  '08})}. \bibinfo{publisher}{ACM}, \bibinfo{address}{New York, NY, USA},
  \bibinfo{pages}{1096--1103}.
\newblock
\showISBNx{978-1-60558-205-4}
\urldef\tempurl%
\url{https://doi.org/10.1145/1390156.1390294}
\showDOI{\tempurl}


\bibitem[\protect\citeauthoryear{Wang, Liu, Cao, Lian, and Liu}{Wang
  et~al\mbox{.}}{2018}]%
        {Wangy2017Attentionbased}
\bibfield{author}{\bibinfo{person}{Shoujin Wang}, \bibinfo{person}{liang Liu},
  \bibinfo{person}{Longbin Cao}, \bibinfo{person}{Defu Lian}, {and}
  \bibinfo{person}{wei Liu}.} \bibinfo{year}{2018}\natexlab{}.
\newblock \showarticletitle{Attention-based Transactional Context Embedding for
  Next-Item Recommendation.}. In \bibinfo{booktitle}{\emph{AAAI}}.
\newblock


\bibitem[\protect\citeauthoryear{Zhang, Dai, Xu, Feng, Wang, Bian, Wang, and
  Liu}{Zhang et~al\mbox{.}}{2014}]%
        {zhang2014sequential}
\bibfield{author}{\bibinfo{person}{Yuyu Zhang}, \bibinfo{person}{Hanjun Dai},
  \bibinfo{person}{Chang Xu}, \bibinfo{person}{Jun Feng},
  \bibinfo{person}{Taifeng Wang}, \bibinfo{person}{Jiang Bian},
  \bibinfo{person}{Bin Wang}, {and} \bibinfo{person}{Tie-Yan Liu}.}
  \bibinfo{year}{2014}\natexlab{}.
\newblock \showarticletitle{Sequential Click Prediction for Sponsored Search
  with Recurrent Neural Networks}.
\newblock   \bibinfo{volume}{2} (\bibinfo{date}{04} \bibinfo{year}{2014}).
\newblock


\bibitem[\protect\citeauthoryear{{Zhou}, {Song}, {Zhu}, {Ma}, {Yan}, {Dai},
  {Zhu}, {Jin}, {Li}, and {Gai}}{{Zhou} et~al\mbox{.}}{2017}]%
        {DIN}
\bibfield{author}{\bibinfo{person}{G. {Zhou}}, \bibinfo{person}{C. {Song}},
  \bibinfo{person}{X. {Zhu}}, \bibinfo{person}{X. {Ma}}, \bibinfo{person}{Y.
  {Yan}}, \bibinfo{person}{X. {Dai}}, \bibinfo{person}{H. {Zhu}},
  \bibinfo{person}{J. {Jin}}, \bibinfo{person}{H. {Li}}, {and}
  \bibinfo{person}{K. {Gai}}.} \bibinfo{year}{2017}\natexlab{}.
\newblock \showarticletitle{{Deep Interest Network for Click-Through Rate
  Prediction}}.
\newblock \bibinfo{journal}{\emph{ArXiv e-prints}} (\bibinfo{date}{June}
  \bibinfo{year}{2017}).
\newblock
\showeprint[arxiv]{stat.ML/1706.06978}


\bibitem[\protect\citeauthoryear{Zhou, Song, Zhu, Ma, Yan, Dai, Zhu, Jin, Li,
  and Gai}{Zhou et~al\mbox{.}}{2017}]%
        {zhou2017deep}
\bibfield{author}{\bibinfo{person}{Guorui Zhou}, \bibinfo{person}{Chengru
  Song}, \bibinfo{person}{Xiaoqiang Zhu}, \bibinfo{person}{Xiao Ma},
  \bibinfo{person}{Yanghui Yan}, \bibinfo{person}{Xingya Dai},
  \bibinfo{person}{Han Zhu}, \bibinfo{person}{Junqi Jin}, \bibinfo{person}{Han
  Li}, {and} \bibinfo{person}{Kun Gai}.} \bibinfo{year}{2017}\natexlab{}.
\newblock \showarticletitle{Deep interest network for click-through rate
  prediction}.
\newblock \bibinfo{journal}{\emph{arXiv preprint arXiv:1706.06978}}
  (\bibinfo{year}{2017}).
\newblock


\end{thebibliography}

\end{document}